\documentclass[conference]{IEEEtran}
\usepackage[draft]{todonotes}
\usepackage{graphicx}
\usepackage{hyperref}
\usepackage[labelformat=simple]{subfig}
\usepackage{amsmath,amssymb} 
\usepackage{color}
\usepackage{multirow}

\ifCLASSINFOpdf
\else
\fi
\begin{document}
%
\title{Localizing and Quantifying Damage in \\Social Media Images}

\author{\IEEEauthorblockN{Xukun Li}
\IEEEauthorblockA{Department of Computer Science\\
Kansas State University\\
Manhattan, Kansas\\
Email: xukun@ksu.edu}\\
\IEEEauthorblockN{Doina Caragea}
\IEEEauthorblockA{Department of Computer Science\\
Kansas State University\\
Manhattan, Kansas\\
Email: dcaragea@ksu.edu} 
\and
\IEEEauthorblockN{Huaiyu Zhang}
\IEEEauthorblockA{Department of Statistics\\
Kansas State University\\
Manhattan, Kansas\\
Email: huaiyu@ksu.edu} \\
\IEEEauthorblockN{Muhammad Imran}
\IEEEauthorblockA{Qatar Computing Research Institute\\
Hamad Bin Khalifa University \\
Doha, Qatar\\
Email: mimran@hbku.edu.qa}

}


%


\maketitle

\begin{abstract}
Traditional post-disaster assessment of damage heavily relies on expensive GIS data, especially remote sensing image data. In recent years, social media has become a rich source of disaster information that may be useful in assessing damage at a lower cost. Such information includes text (e.g., tweets) or images posted by eyewitnesses of a disaster. Most of the existing research explores the use of text in identifying situational awareness information useful for disaster response teams. The use of social media images to assess disaster damage is limited. In this paper, we propose a novel approach, based on convolutional neural networks and class activation maps, to locate damage in a disaster image and to quantify the degree of the damage. Our proposed approach enables the use of  social network images for post-disaster damage assessment, and provides an inexpensive and feasible alternative to the more expensive GIS approach.
\end{abstract}


%
\IEEEpeerreviewmaketitle

\section{Introduction}

Fast detection of damaged areas after an emergency event can  inform responders and aid agencies, support logistics involved in relief operations,  accelerate real-time response, and guide the allocation of resources. Most of the existing studies on detecting and assessing disaster damage rely heavily on macro-level images, such as remote sensing imageries \cite{xie2016crowdsourcing,gueguen2015large}, \cite{fan2017quantifying}, or imageries transmitted by unmanned aerial vehicles \cite{attari2017nazr}. Collection and analysis of macro-level images require costly resources, including expensive equipment, complex data processing tools, and also good weather conditions. To benefit the response teams, the macro-level images have to be collected and analyzed  very fast, which is not always possible with traditional collection and analysis methods. 

With the growth of social media platforms in recent years, real-time disaster-related information is readily available, in the form of network activity (e.g., number of active users, number of messages posted), text (e.g., tweets), and images posted by eyewitnesses of disasters on platforms such as Twitter, Facebook, Instagram or Flicker. Many studies have shown the utility of social media information for disaster management and response teams. For example,  the analysis of text data (e.g., tweets from Twitter) has received significant attention in recent works \cite{guan2014using}, \cite{imran2015processing}, \cite{kryvasheyeu2016rapid}, \cite{hongmin:JCCM}, \cite{yuan2018feasibility}. 
However, social media images, while very informative \cite{Bica:2017}, have not been extensively used to aid disaster response, primarily due to the complexity of information extraction from (noisy) images, as compared to information extraction from text. 

By contrast with macro-level images, social media images have higher ``resolution'', in the sense that they can provide detailed on-site information from the perspective of the eyewitnesses of the disaster \cite{Bica:2017}. Thus, social media images can serve as an ancillary yet rich source of visual information in disaster damage assessment. Pioneering works with focus on the utility of social media images in disaster response  include \cite{alam2017image4act}, \cite{nguyen2017damage}, where the goal is to use convolutional neural networks (CNN) to assess the severity of the damage (specifically, to classify social media images based on the degree of the damage as: {\it severe}, {\it mild}, and {\it none}). 

Due to ground-breaking developments in computer vision, many image analysis tasks have become possible. Disaster management and response teams can benefit from novel image analyses that can produce quantitative assessments of damage, and inform the relief operations with respect to priority areas. In this context, it would be useful to locate damage areas in social media images (when images contain damage), and subsequently use the identified damage areas to assess the damage severity on a continuous scale. Possible approaches for localizing damage in social media images include object detection \cite{cha2017deep,maeda2018road} and image segmentation \cite{vetrivel2017disaster}. Object detection can be conducted by classifying some specific regions in an image as containing damage or not. For example, Cha et al. \cite{cha2017deep} used a convolution neural network (CNN) to classify small image regions (with $256 \times 256$ pixel resolutions) as containing concrete crack damage or not. 
Maeda et al. \cite{maeda2018road} used a state-of-the-art object detection approach, called Single Shot MultiBox Detector (SSD) \cite{LiuAESR15}, to detect several types of road damage. Image segmentation 
has been used in \cite{vetrivel2017disaster} to detect building damage based on high resolution aerial images. 

Regardless of the method used, object detection or image segmentation, existing approaches for localizing damage first identify objects (i.e., potential damage regions) and subsequently classify the objects as {\it damage}  (sometimes, {\it severe} or {\it mild}) or {\it no damage}. Thus, there is a conceptual mismatch in the way existing approaches are used, given that damage is generally regarded as a high-level concept rather than a well-defined object. By first identifying objects and then assigning discrete hard-labels to them, existing approaches produce a clear-cut boundary for the damaged areas, although a smooth boundary would be more appropriate. 

{\bf Contributions:} Inspired by the technique called Class Activation Mapping (CAM) \cite{zhou2016learning}, we propose a novel approach, called Damage Detection Map (DDM), to generate a smooth damage heatmap for an image. Our approach adopts the gradient-weighted CAM \cite{selvaraju2016grad} technique to localize the area in an image which contributes to the damage class. 
Based on the damage heatmap, we also propose a new quantitative measure, called Damage Assessment Value (DAV), to quantify the severity of damage on a continuous scale. 
Our approach makes the disaster damage localization possible, and thus extends the use of social media images in disaster assessment.

The rest of this paper is organized as follows: we  describe the proposed approaches for generating  DDM heatmaps, and computing DAV scores in Section \ref{method}. We describe the experimental setup and results in Section \ref{results}. We discuss related work in Section \ref{related}, and conclude the paper in Section \ref{conc}.

\section{Proposed Approach}
\label{method}
\begin{figure*}%
    \centering
	\includegraphics[width=\textwidth]{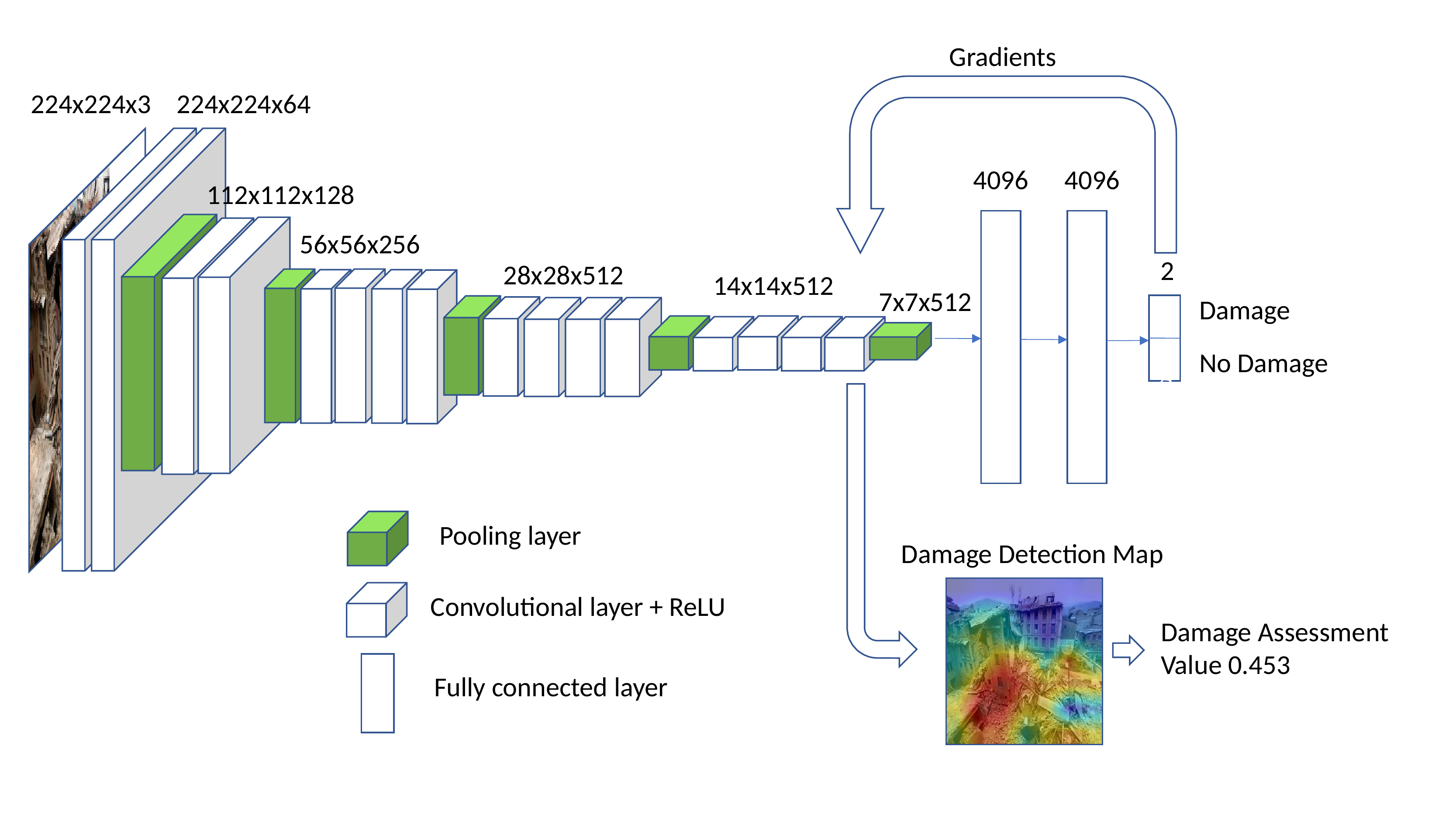}%
    \caption{Overview of the proposed approach. }%
    \label{fig:model}%
\end{figure*}

Our approach generates a Damage Detection Map, which visualizes the damage area for a given image, and a score DAV, which quantifies the severity of the damage. The main components of our approach, shown in Fig. \ref{fig:model}, are the following: 1) a CNN that classifies images into two classes, {\it damage} or {\it no damage}; 2) a class activation mapping, which generates the DDM map by weighting the last convolutional layer of the CNN model; 3) finally, the damage severity score computed by averaging the values in the map. The details for the three components of our approach are provided in what follows.

\subsection{Convolutional Neural Networks}
\label{CNN}
Convolutional neural networks \cite{lecun1989backpropagation} have been used successfully for many image analysis tasks \cite{lecun2015deep, goodfellow2016deep}. The ImageNet annual competition (where a dataset with 1.2 million images in 1000 categories is provided to participants) has led to several popular architectures, including AlexNet \cite{krizhevsky2012imagenet}, VGG19 \cite{simonyan2014very}, ResNet \cite{he2016deep} and Inception \cite{inception}. We choose VGG19 as the architecture for our CNN model, as VGG19 has good classification accuracy and it is relatively simpler compared to ResNet and Inception. Furthermore, the pre-trained model are available for VGG19 and it is easy to fine tune them for different classification problems. 

VGG19 \cite{simonyan2014very} contains 16 convolutional layers (with 5 pooling layers) and 3 fully connected layers. Each convolution layer is equipped with a non-linear ReLU activation \cite{krizhevsky2012imagenet}. The convolutional layers can be seen as feature extraction layers, where each successive layer detects predictive image features (i.e., image fragments that correspond to edges, corners, textures, etc.) at a more abstract level than the previous layer. 

As can be seen in Fig. \ref{fig:model}, the size of the input to the convolutional layers is reduced by a factor of 2 through max-pooling layers, but not all convolution layers are followed by a max-pooling layer.  After every max-pooling layer, the width of the convolution layer (i.e., number of filters used) increases by a factor of 2. After the last max-pooling layer, there are two fully connected layers with dimension 4096, and another fully connected layer whose neurons correspond to the categories to be assigned to the input image. The last layer of the standard VGG19 model has dimension 1000 because VGG19 was originally trained on a dataset with 1000 categories. However, as we are interested in using VGG19 to classify images in two categories, {\it damage} and {\it no damage}, we change the dimension of the last fully connected layer from 1000 to 2.  Overall, the model includes more than 130 million parameters, and it takes a significant amount of time (and a large number of images) to train it accurately \cite{simonyan2014very}. However, the model parameters are highly transferable to other image classification problems \cite{yosinski2014transferable}.  Thus, to avoid the need for a large number of images, we initialize our model with the pre-trained VGG19 model (except for the last fully connected layer), and fine tune it using disaster-related images. 
More specifically, given a training image $x$ with label $y$ represented as a one-hot vector (e.g., if the label is {\it damage}, then $y=[1, 0]$, otherwise $y=[0, 1]$), all the parameters $\theta$ will be updated by:

\begin{equation}\label{update}
\theta \longleftarrow \theta -\mu \frac{\partial \mathcal{L}(y, \text{CNN}(x))}{\partial \theta},
\end{equation}
where $\mu$ is learning rate, $\mathcal{L}$ is the cross-entropy loss, and $\text{CNN}(x)$ is the output of the CNN given input $x$.

\subsection{Damage Detection Map}
\label{DDM}
Our proposed Damage Detection Map (DDM) is inspired by the Gradient-weighted Class Activation Mapping (GCAM) \cite{selvaraju2016grad}. In a general classification problem, for an input image and a trained CNN model, GCAM makes use of the gradients of a target category to compute a category-specific weight for each feature map of a convolution layer. The weights are  used to aggregate the feature maps of the final convolutional layer, under the assumption that the last level captures the best trade-off between high-level semantic features and spatial information. The resulting maps can be used to identify the discriminative regions for the target category (which explain the CNN model's prediction), and implicitly to localize the category in the input image. Thus, GCAM can be seen as a weakly supervised approach, which can localize a category in an image based only on global image labels \cite{selvaraju2016grad}. Furthermore, the GCAM localized categories or objects (shown using heatmaps) have soft boundaries, and can be used to gain both insight and trust into the model. This makes GCAM particularly attractive for the problem of localizing damage in disaster images, assuming that only coarse labeling of images as {\it damage} or {\it not damage} is available for training. The heatmaps showing categories of interest using soft-boundaries are very appropriate for localizing damage, as damage boundaries are inherently soft. Moreover, the heatmaps that explain the model's predictions can be used to gain the trust of disaster management teams, and thus increase the usability of social media images in disaster response and recovery.

Given the above introduction to GCAM and motivation for use in the context of disaster damage localization, we formally describe the GCAM approach \cite{selvaraju2016grad} in remaining of this subsection. 
Consider a $14\times14$ matrix $S$, such that 
\begin{equation}\label{s}s_{i,j} = \displaystyle
\text{ReLU}
(\sum_kw_kf^k_{(i, j)}
)
\end{equation}where $w_k$ is a gradient-based weight parameter (defined in Equation \ref{w}) corresponding to each feature map $f^k$ in the last convolutional layer (of dimension $14\times 14\times 512$), and $f^k_{(i, j)}$ represents the value at location ${(i, j)}$ in the $k$-th feature map, for $i = 1,\ldots, 14$, $j= 1,\ldots, 14$, $k = 1,\ldots, 512$.  The ReLU function is used to cancel the effect of the negative values, while emphasizing the effect of the positive values. Given an input image $x$ and the output of the CNN for the damage class $y_D$ (just before the softmax function is applied), the weights $w_k$ are determined as the sum of the gradients of output $y_D$ with respect to $f^k_{(i, j)}$, for all $i,j$. Specifically: 
\begin{equation}\label{w}
w_k = \dfrac{1}{14\times 14}
\sum_{i,j}\frac{\partial y_D}{\partial f^k_{(i,j)}}
\end{equation}
The feature maps $f^k$ in Equations (\ref{w}) and (\ref{s}) are in the last convolutional layer, as this layer generally shows a good trade-off between high-level features and spatial information in the original image. Our final goal is to localize damage in disaster images, in other words to find regions that are discriminative for the damage class. Intuitively, the gradient with respect to a neuron in the final convolutional layer represents the contribution of the neuron to the class label. The procedure described above produces a $14\times14$ matrix $S$ (and correspondingly a $14\times14$ image). Using an interpolation technique, we further resize $S$ to $S_C$ to match the original dimensions of the image. The heatmap showing the $S_C$ values is the final Damage Detection Map. 

\subsection{Measuring the damage severity}
\label{DAV}
When a disaster occurs, eyewitnesses of the disaster will produce a huge number of images in a short period of time. An important goal of disaster assessment is to extract and concisely summarize the information contained in the images posted by eyewitnesses. To meet this demand, we propose to use a damage assessment value (DAV) derived from the DDM heatmap to represent the damage severity for each image. 


The disaster damage map uses numerical values to measure the intensity of each pixel of the image (the higher the intensity, the more severe the damage), and can be represented as a heatmap. We take the average over all the numerical values in the heatmap of a given image, and use the resulting value as an overall score for the severity of the damage. Formally, we define the damage assessment value (DAV) as:
\begin{equation}
DAV =  \frac{1}{14\times 14}\sum_{i,j} s_{i,j}
\end{equation}
where $s_{i,j}$ are the elements of the $S$ matrix defined in Equation (\ref{s}), and   $14\times14$ is the dimension of the matrix $S$. 


\begin{figure*}%
    \centering
    \subfloat[(1) Damage]{{\includegraphics[width=1.4in]{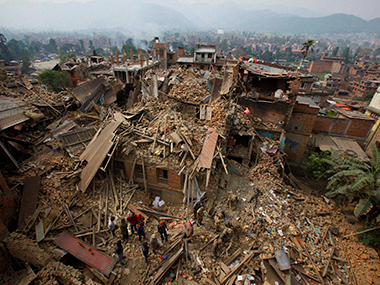} }}%
    \subfloat[(2) Damage]{{\includegraphics[width=1.4in]{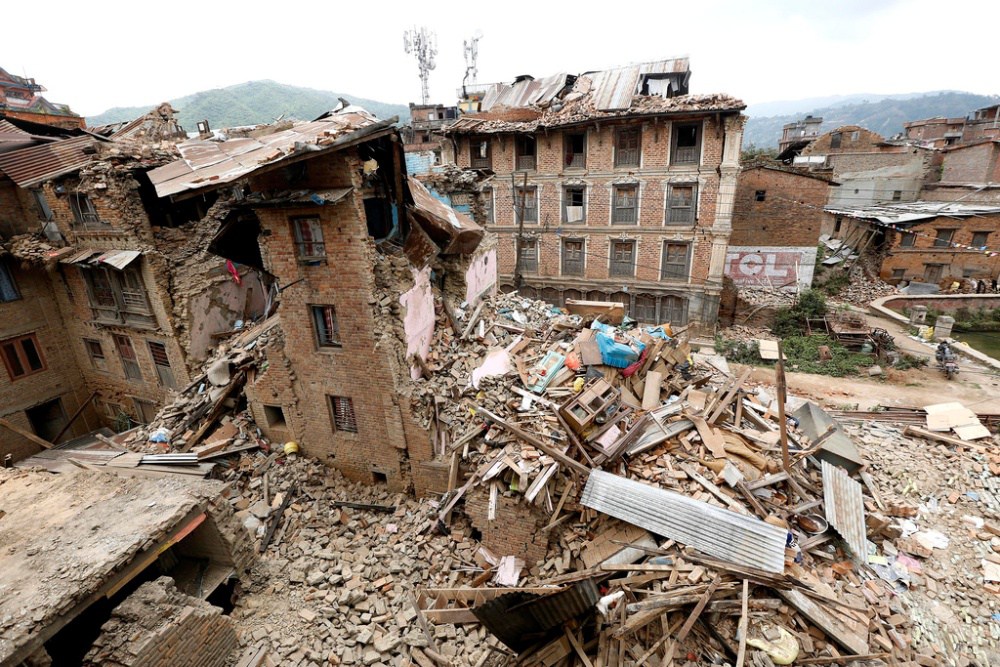} }}%
    \subfloat[(3) Damage]{{\includegraphics[width=1.4in]{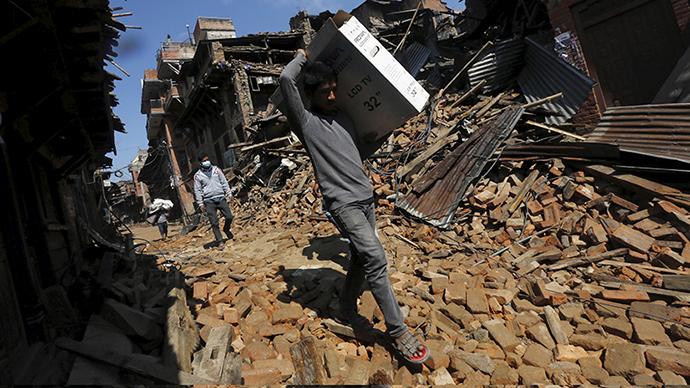} }}%
    \subfloat[(4) Damage]{{\includegraphics[width=1.4in]{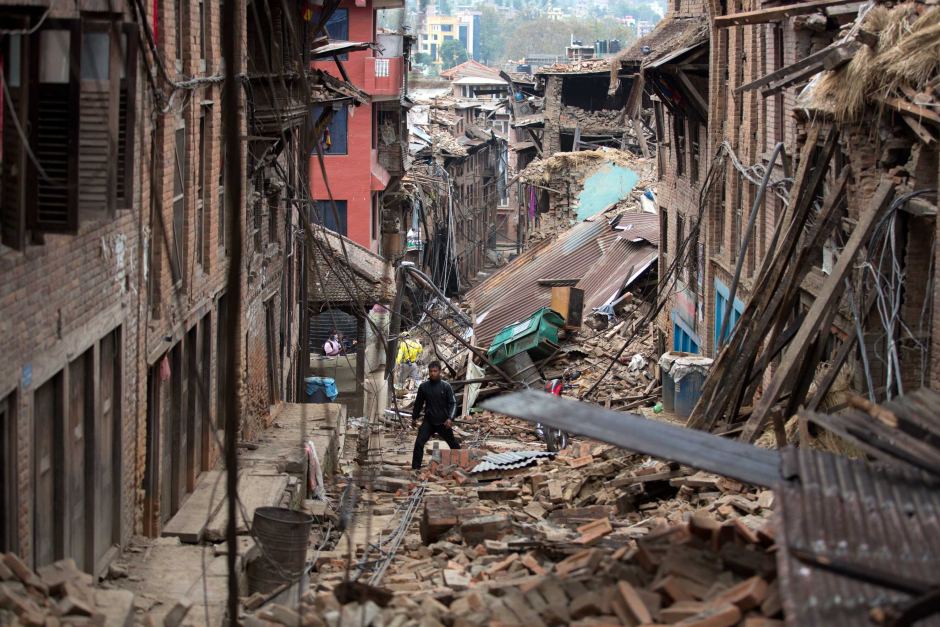} }}
    \qquad
    \subfloat[(1) DAV = 0.413 ]{{\includegraphics[width=1.4in]{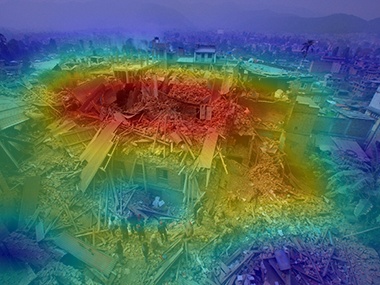} }}%
    \subfloat[(2) DAV = 0.453]{{\includegraphics[width=1.4in]{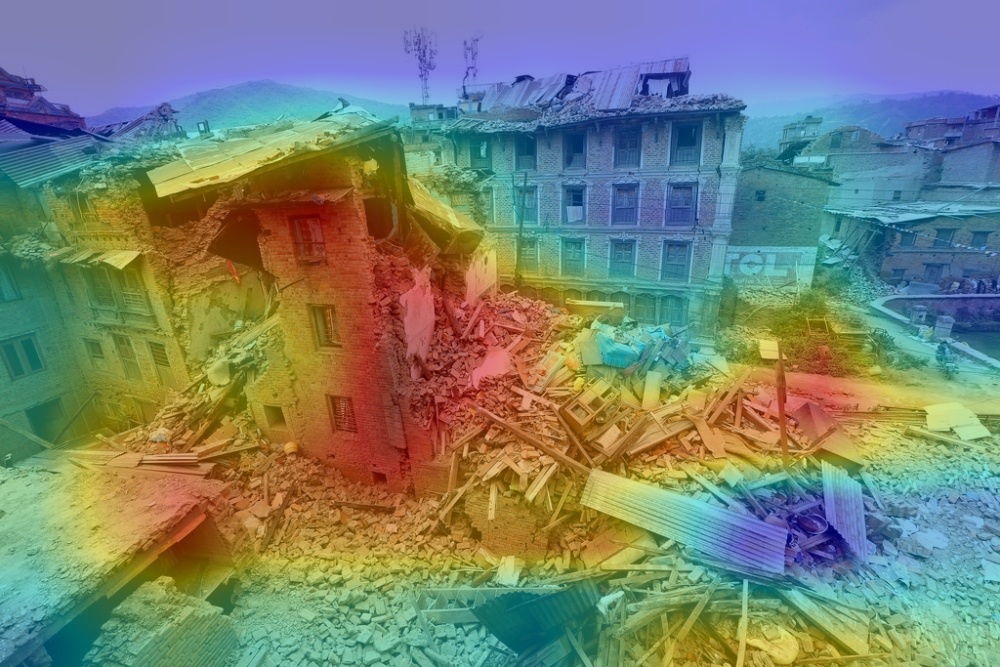} }}%
    \subfloat[(3) DAV = 0.423]{{\includegraphics[width=1.4in]{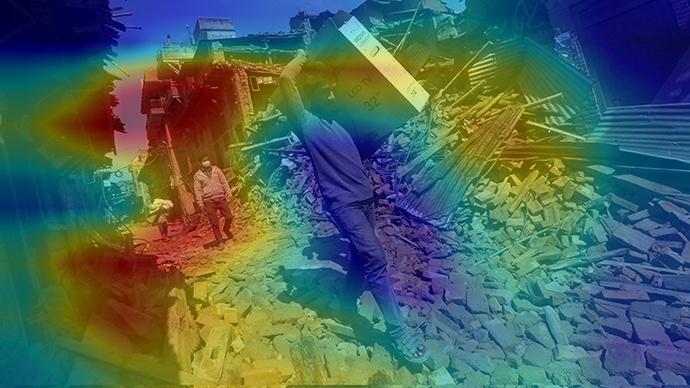} }}%
    \subfloat[(4) DAV = 0.385]{{\includegraphics[width=1.4in]{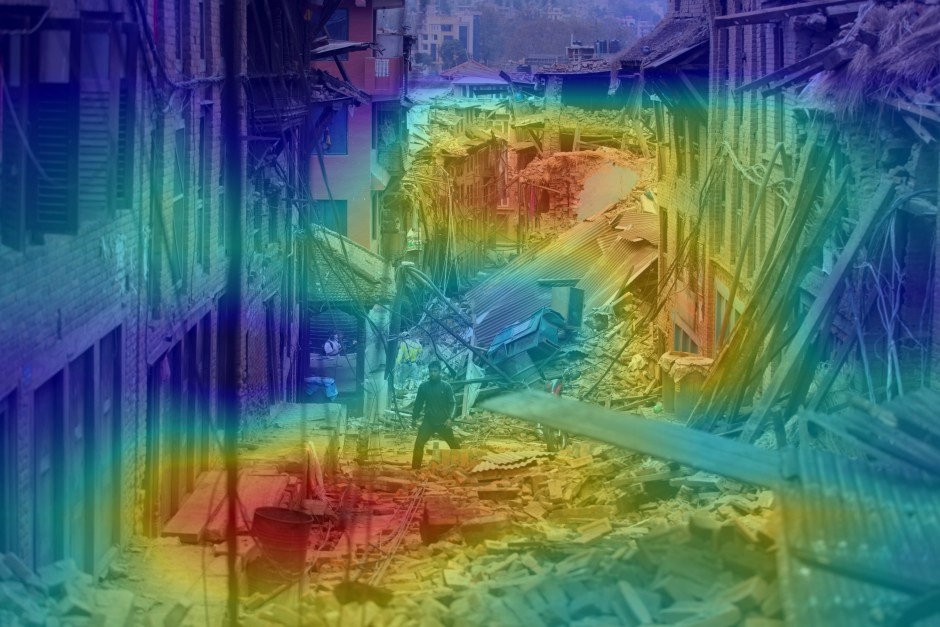} }}
    \qquad
    \subfloat[(5) No damage]{{\includegraphics[width=1.4in]{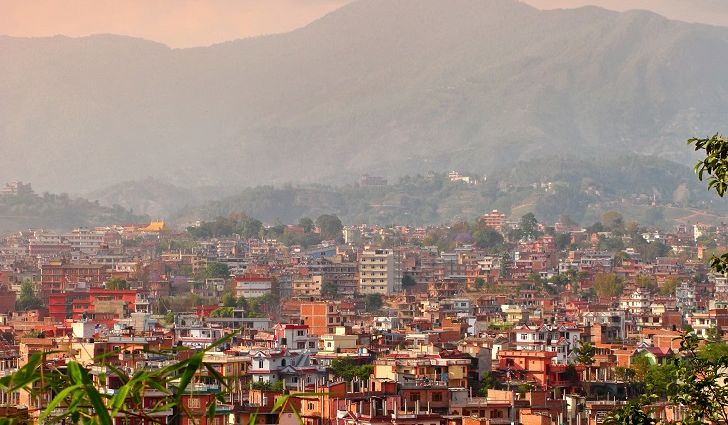} }}%
    \subfloat[(6) No damage]{{\includegraphics[width=1.4in]{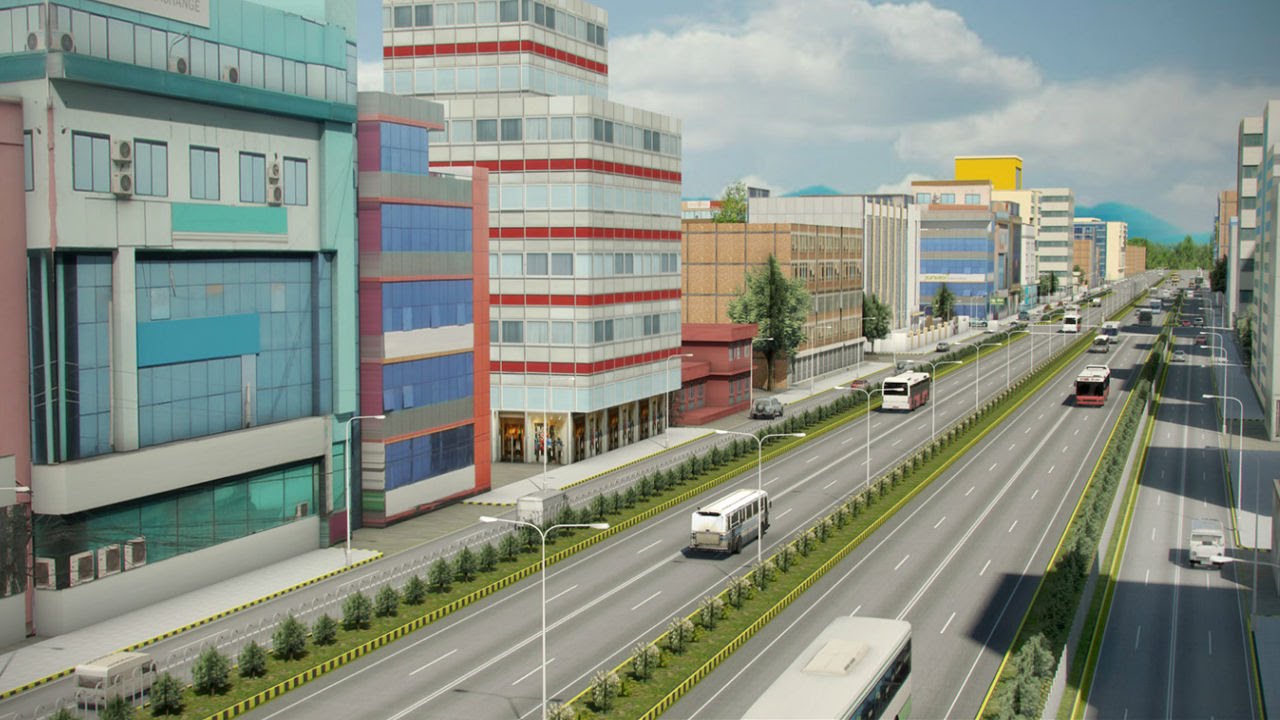} }}%
    \subfloat[(7) No damage]{{\includegraphics[width=1.4in]{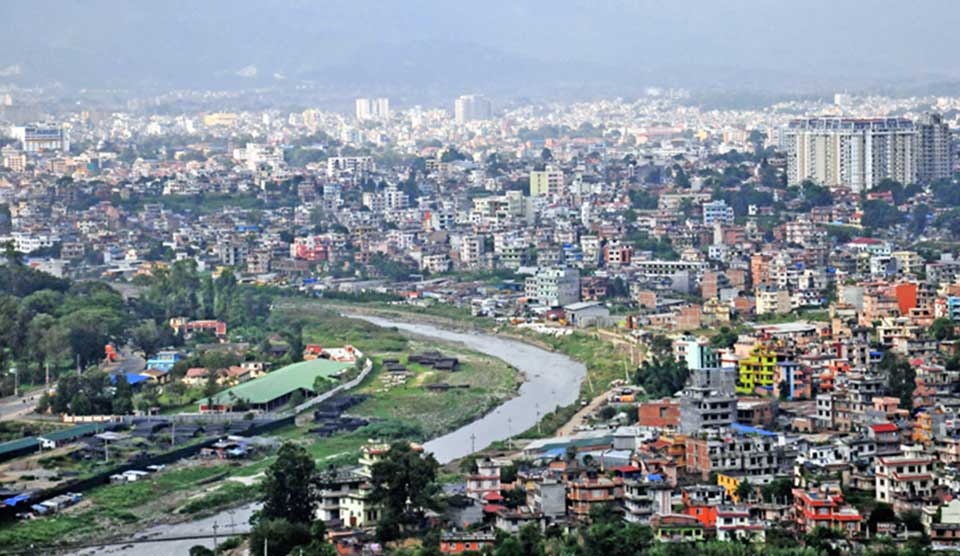} }}%
    \subfloat[(8) No damage]{{\includegraphics[width=1.4in]{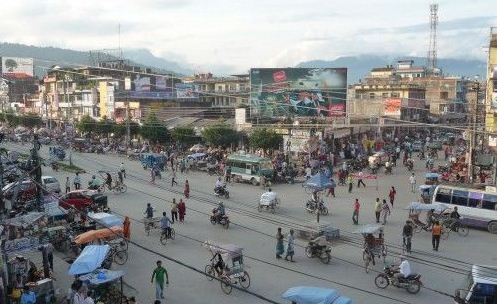} }}
    \qquad
    \subfloat[(5) DAV = 0.01]{{\includegraphics[width=1.4in]{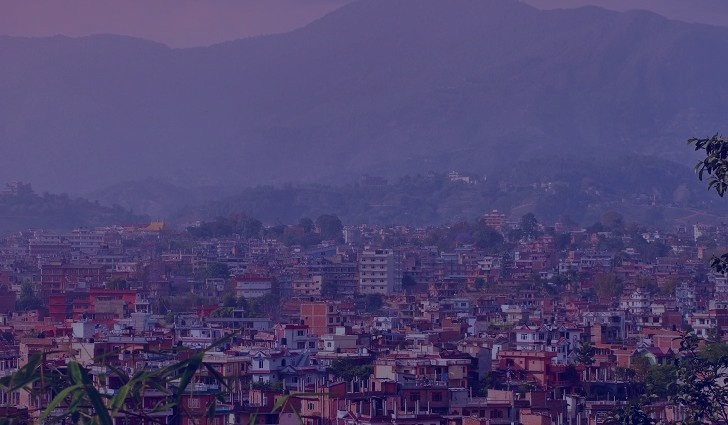} }}%
    \subfloat[(6) DAV = 0.028]{{\includegraphics[width=1.4in]{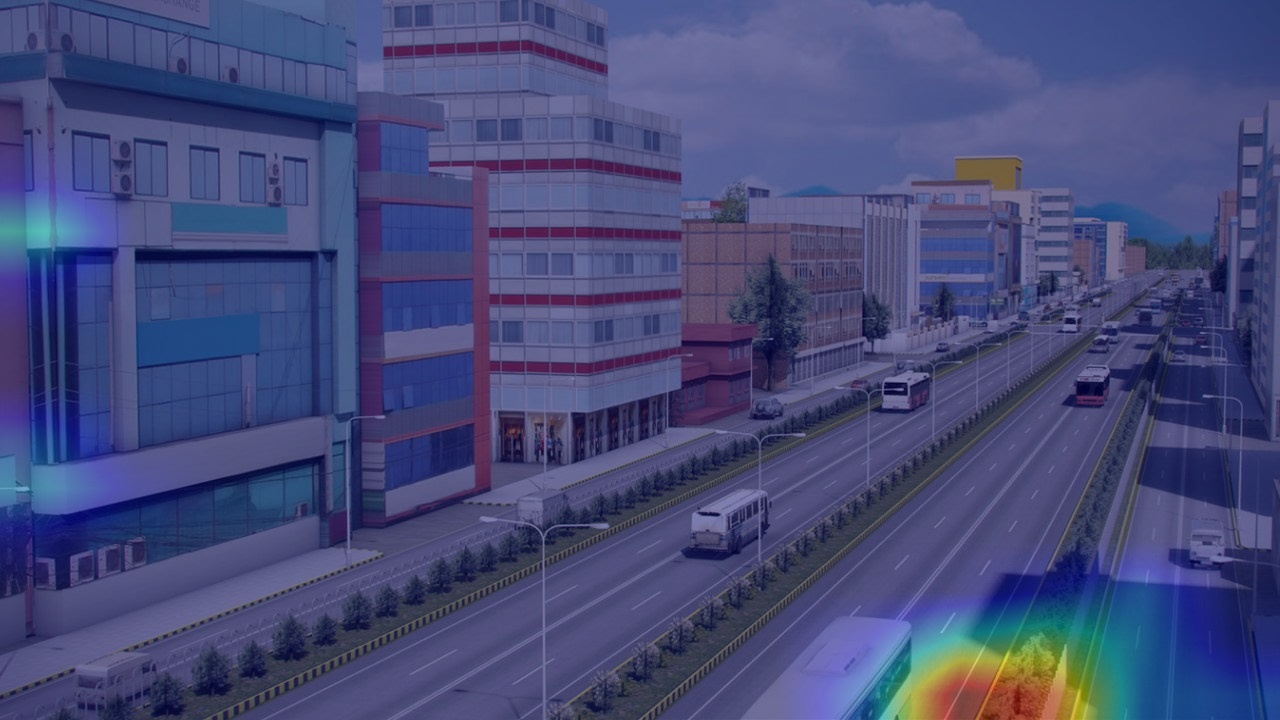} }}%
    \subfloat[(7) DAV = 0.0]{{\includegraphics[width=1.4in]{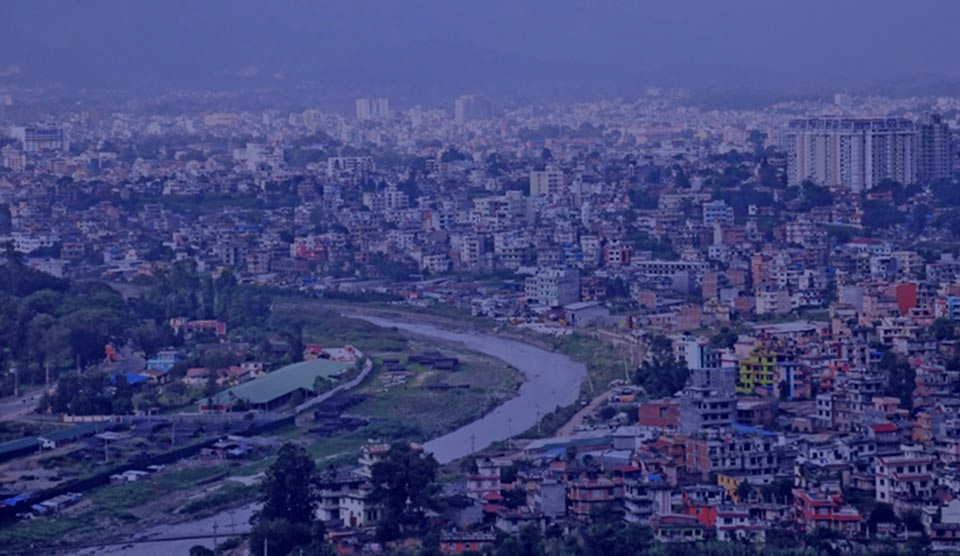} }}%
    \subfloat[(8) DAV = 0.033]{{\includegraphics[width=1.4in]{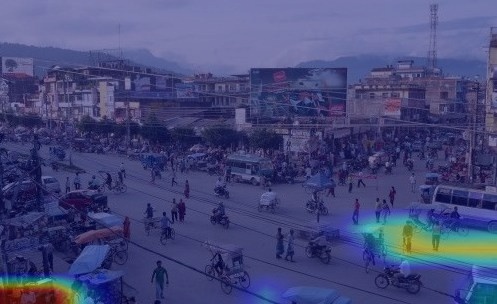} }}
    \caption{Examples of images in our dataset, and their corresponding DDM heatmaps and DAV scores. The first row shows examples of images in the {\it damage} class, followed by their corresponding DDM heatmaps and the DAV scores. Similarly, the third row shows examples of images in the {\it no damage} class, followed by their corresponding DDM heatmaps and DAV scores. }%
    \label{fig:data_example}%
\end{figure*}

\section{Experimental Results}
\label{results}
We perform a series of experiments to evaluate the performance of our proposed method. The  experiments are designed to answer the following questions: (i) can the Damage Detection Map accurately locate the damage areas, (ii) can the DAV score provide a reliable measure for damage severity. 

\subsection{Experimental Setting}

\subsubsection{Datasets} We used two datasets in our experiments. The first dataset is assembled using Google image search engine. Specifically, we performed two searches: first, we used `nepal', `building' and `damage' as keywords, and crawled $308$ {\it damage} images from the result; second, we  used `nepal' and `city' as keywords, and crawled $311$ {\it no damage} images. The keyword `nepal' was used in both searches to ensure that the two sets of images have similar scene, while the keywords `building damage' and `city' were used to bias the search towards  {\it damage} and {\it no damage} images, respectively. We show some sample {\it damage} and {\it no damage} images from the Google dataset in Fig. \ref{fig:data_example} in the first and third rows, respectively. 

The second dataset used in our evaluation was previously used in \cite{nguyen2017damage}, and consists of social media images posted during four different disaster events: Nepal Earthquake, Ecuador Earthquake, Ruby Typhoon, and Matthew Hurricane. For each disaster, the dataset contains images in three categories, representing three levels of damage: severe, mild, and none. Table \ref{table:data_summary} provides the class distribution for each disaster dataset. 

\begin{table}
\centering
\caption{Social media image dataset consisting of images from four disaster events. Images are labeled as {\it Severe}, {\it Mild}, or {\it None}. The number of images in each class, and the total number of images for each disaster are shown. }
\label{table:data_summary}
\begin{tabular}{|l|r|r|r|r|}
\hline
Disaster   & Severe  & Mild & None & Total \\ \hline
Nepal Earthquake & 5303  & 1767    & 11226   & 18296  \\ \hline
Ecuador Earthquake   & 785  & 83 & 886  & 1754  \\ \hline
Ruby  Typhoon   & 76  & 325  & 463  & 864  \\ \hline
Matthew Hurricane & 97  & 89  & 130  & 316   \\ \hline
\end{tabular}
\end{table}

In our experiments, each dataset is randomly split into a training set ($80\%$) and a test set ($20\%$). 
We will use the proposed approach to learn a CNN that discriminates between {\it damage} (including {\it severe} and {\it mild}) and {\it no damage} images, while also localizing the damage and identifying image features discriminative for the {\it damage} class. 

\subsubsection{Hyper-parameters} We used TensorFlow's GradientDescentOptimizer to train the model using mini-batch gradient descent on a GeForce GTX 1070 graphic card. Based on preliminary experimentation with the Ecuador Earthquake and Ruby Typhoon datasets, we chose to use a learning rate of 0.001 and a batch size of 32 images in all our experiments. Furthermore, we used the dropout technique with a rate of 0.5 to prevent overfitting. {The code for the VGG19 model was adapted from \url{https://github.com/machrisaa/tensorflow-vgg}}.

\subsection{Damage Detection Map Evaluation}

We first use the Google dataset, where the {\it damage} images are easier to discriminate from the {\it no damage} images, to evaluate the Damage Detection Map (DDM) approach (intuitively the better the CNN classifier, the better the DDM map). Specifically, we train a CNN model on the Google dataset as described in Section \ref{CNN}. The training accuracy of the CNN model is $100\%$, and the test accuracy is $95.5\%$. Subsequently, we compute the DDM heatmaps as described in Section \ref{DDM}. The DDM heatmaps corresponding to the sample {\it damage} and {\it no damage} images are shown in Fig. \ref{fig:data_example}, in the second and fourth rows, respectively. As can be seen, the regions with high intensity DDM  generally correspond to damage.

To answer our first research question, specifically to understand if the DDM can accurately locate the damage areas, we will evaluate the localization capability of DDM by using the Intersection-Over-Union (IOU) measure, which is frequently used to evaluate object detection techniques \cite{Everingham:2010}. The IOU measure is defined as follows:
\begin{equation}
\text{IOU} = \frac{\text{Area of Overlap}}{\text{Area of Union}}
\end{equation}
where the `Area of Overlap' and `Area of Union' are computed with respect to a ground truth image, where damage is manually marked. IOU takes values in $[0,1]$. If the IOU value for a pair of GCAM damage marked image and the corresponding ground truth image is large, then the marked damage areas in the two images are similar, and thus the GCAM approach performs well in terms of damage localization. 

We randomly select 10 images from the  Google test dataset for manual annotation, and use the annotations to evaluate damage localization using IOU scores. 
The scores are computed by comparing automatically DDM localized damage with manually marked damage in the corresponding images.
As damage is not an object, heatmaps with smooth boundaries are preferable to bounding boxes when localizing damage. However, human annotators cannot provide precise heatmaps, unless they have professional knowledge of disaster damage, in which case their annotation would be very expensive. To  reduce the cost, generally human annotators will simply mark the regions of an image that contain damage (resulting in a binary included/not included representation).  We used the tool LabelMe   (\url{https://github.com/wkentaro/labelme}) to mark the damage.

To compare heatmaps with images marked by annotators in terms of IOU values, we transform the heatmaps to a binary representation as follows: we determine the maximum value in $S_C$ and use 20\% of the maximum value as a cutoff value for including a region in the disaster damage ``object'' or not \cite{zhou2016learning}.
In other words, only the regions in DDM with values larger than the cutoff value will be part of the  localized damage. In the resulting transformed image (as well as in the human annotated images), the damage pixels have value 255, while no damage pixels have value 0. 
We show the result of the IOU evaluation on the sample consisting of 10 Google images, by comparison with annotations from two independent annotators A and B, in Table \ref{table:iou}. To understand the difficulty of the task of identifying damage in a picture, we also compute IOU values that show the agreement between annotators A and B. 

\begin{figure*}[t]
    \centering
    
            \subfloat[(a) Original 1]{{\includegraphics[width=1.30in]{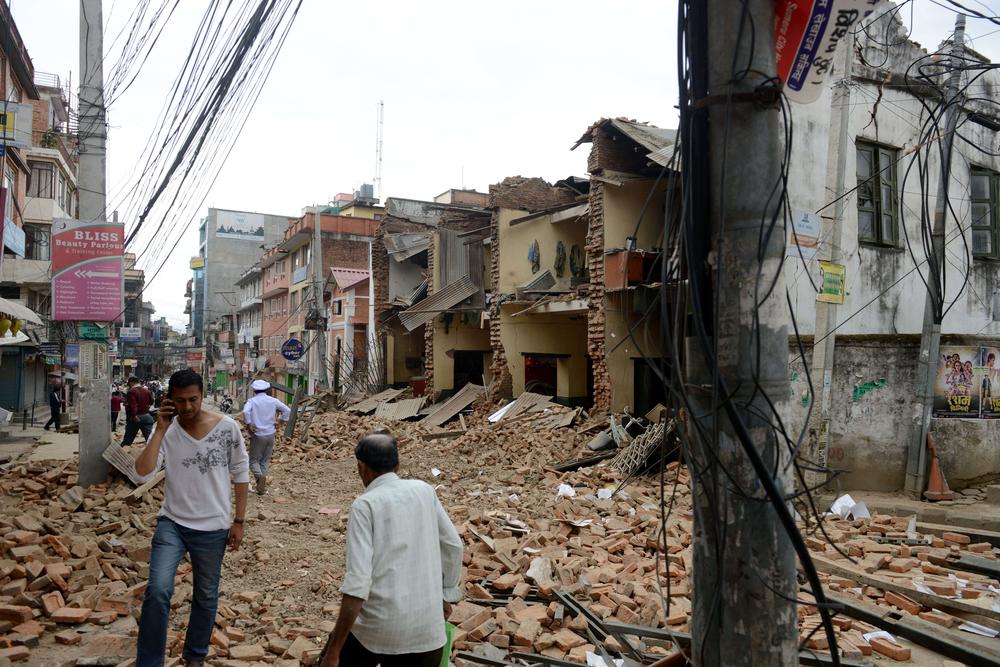} }}%
    \subfloat[(b) DDM]{{\includegraphics[width=1.30in]{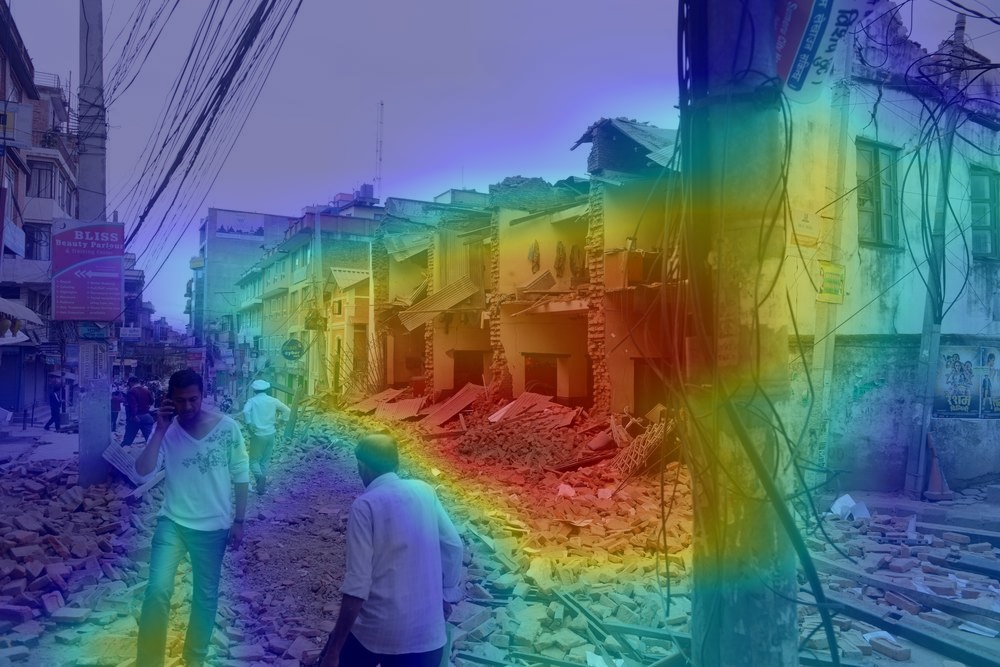} }}%
    \subfloat[(c) DDM+Threshold]{{\includegraphics[width=1.30in]{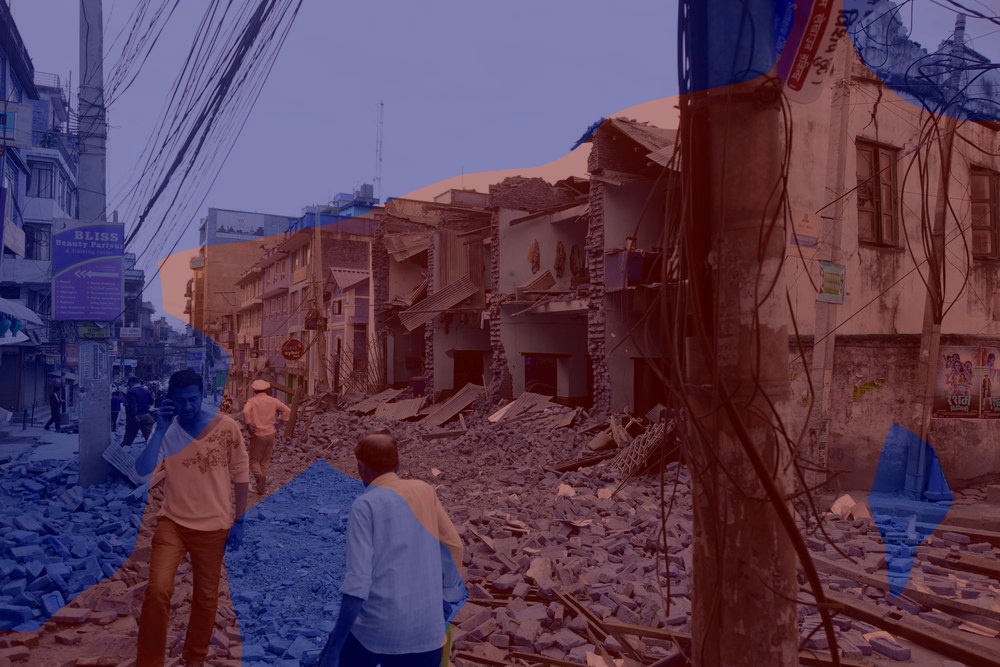} }}%
    \subfloat[(d) Annotator A]{{\includegraphics[width=1.30in]{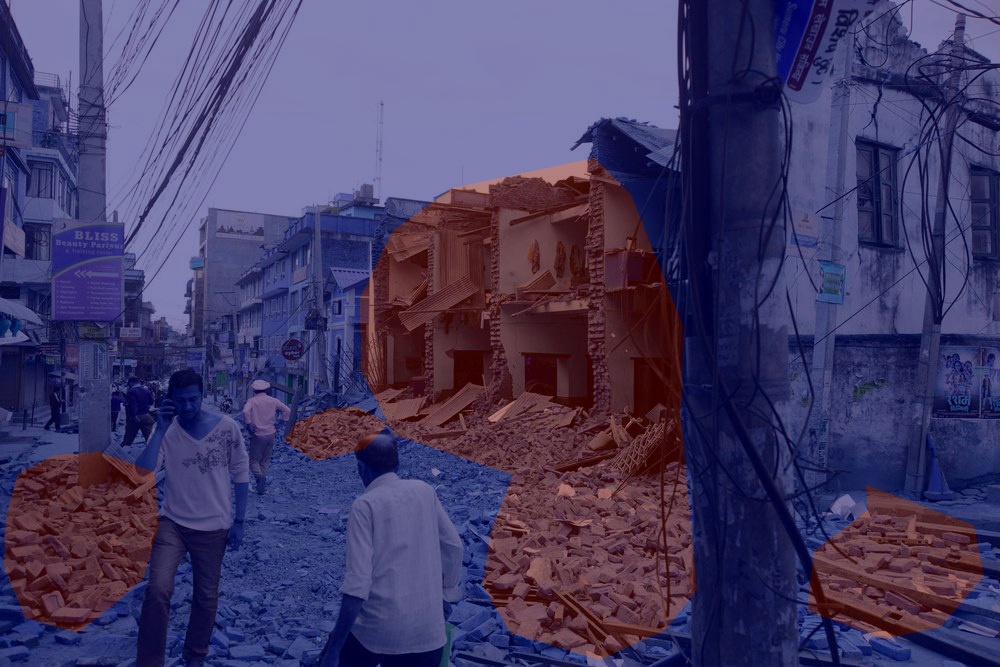} }}%
    \subfloat[(e) Annotator B]{{\includegraphics[width=1.30in]{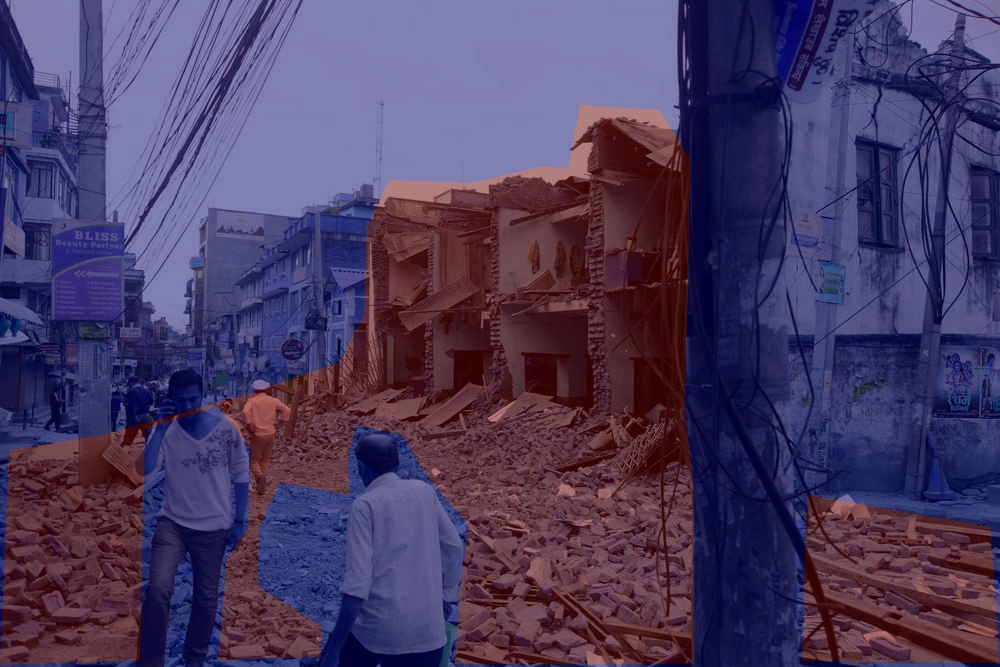} }}
      \qquad 
    \subfloat[(a) Original 5]{{\includegraphics[width=1.30in]{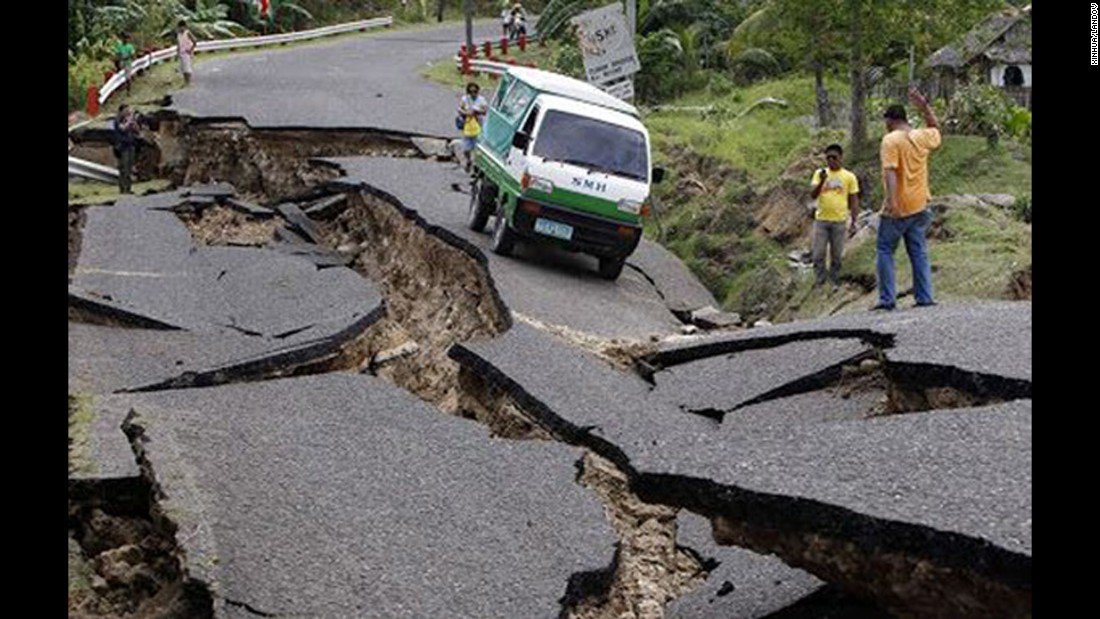} }}%
    \subfloat[(b) DDM]{{\includegraphics[width=1.30in]{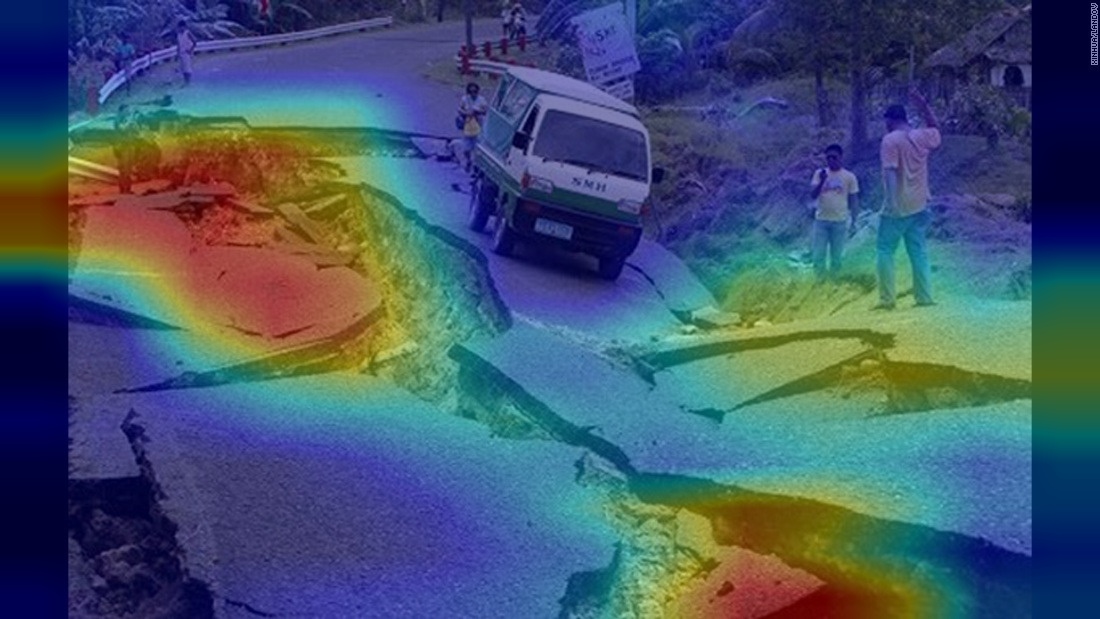} }}%
    \subfloat[(c) DDM+Threshold]{{\includegraphics[width=1.30in]{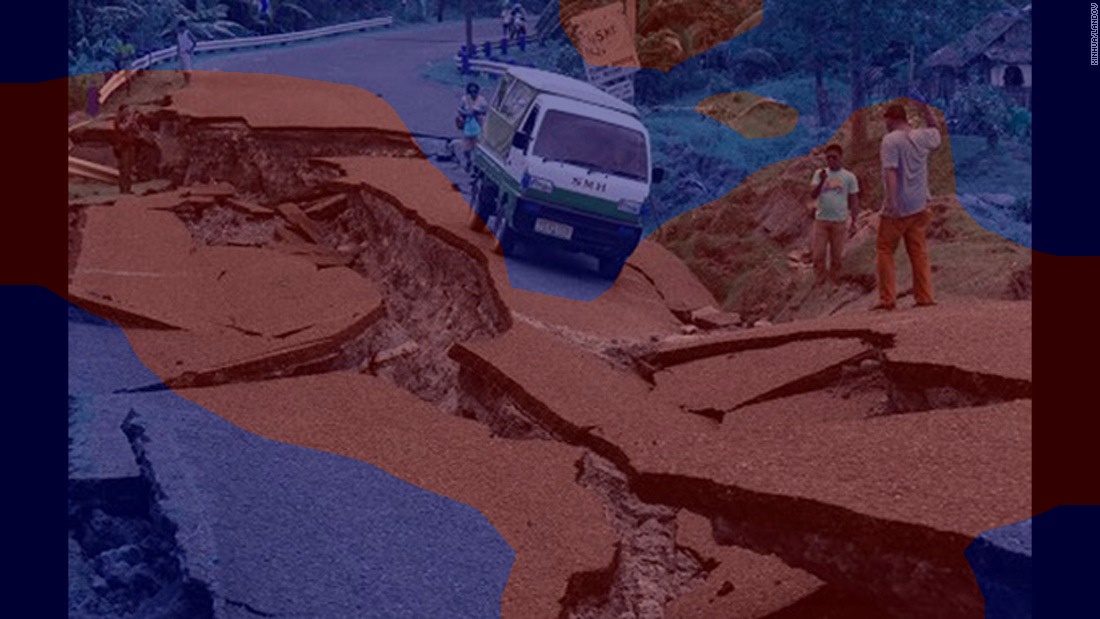} }}%
    \subfloat[(d) Annotator A]{{\includegraphics[width=1.30in]{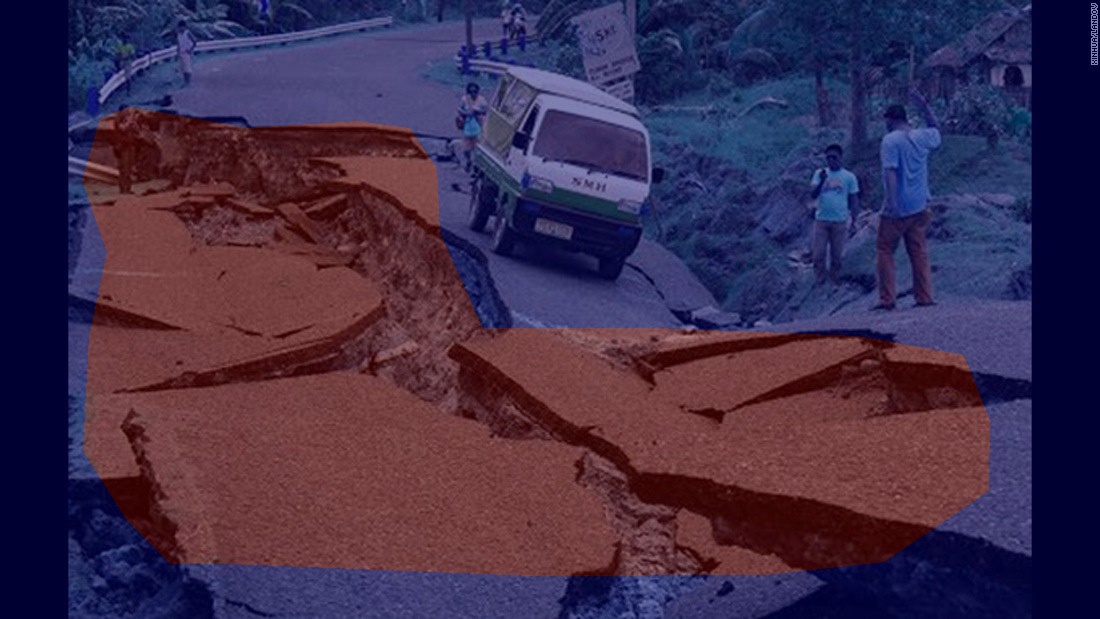} }}%
    \subfloat[(e) Annotator B]{{\includegraphics[width=1.30in]{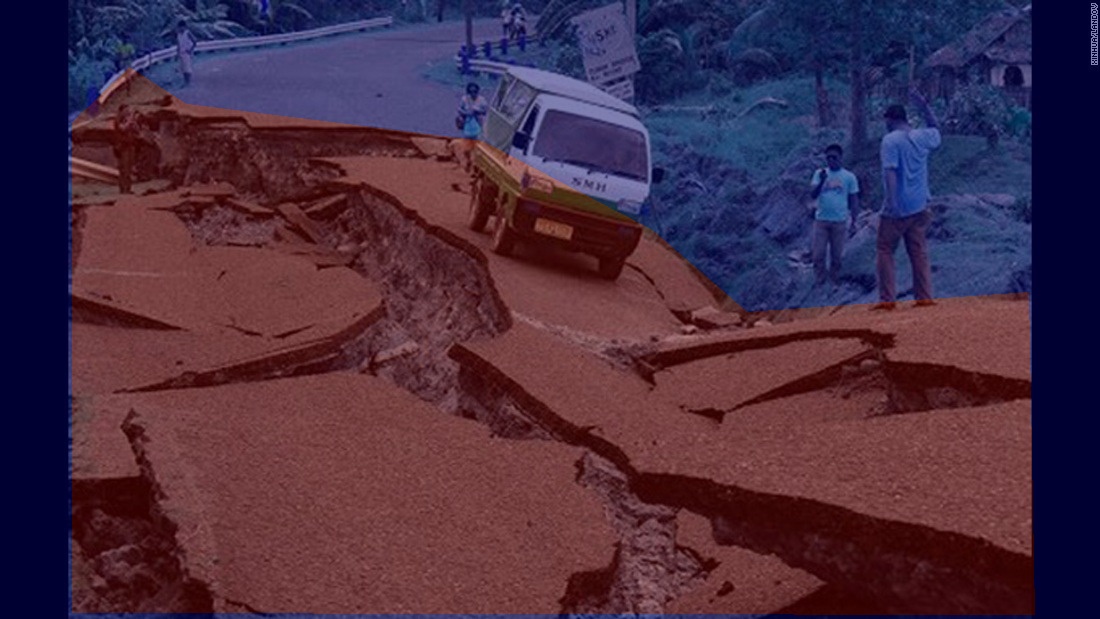} }}
          \qquad  
      \subfloat[(a) Original 7]{{\includegraphics[width=1.30in]{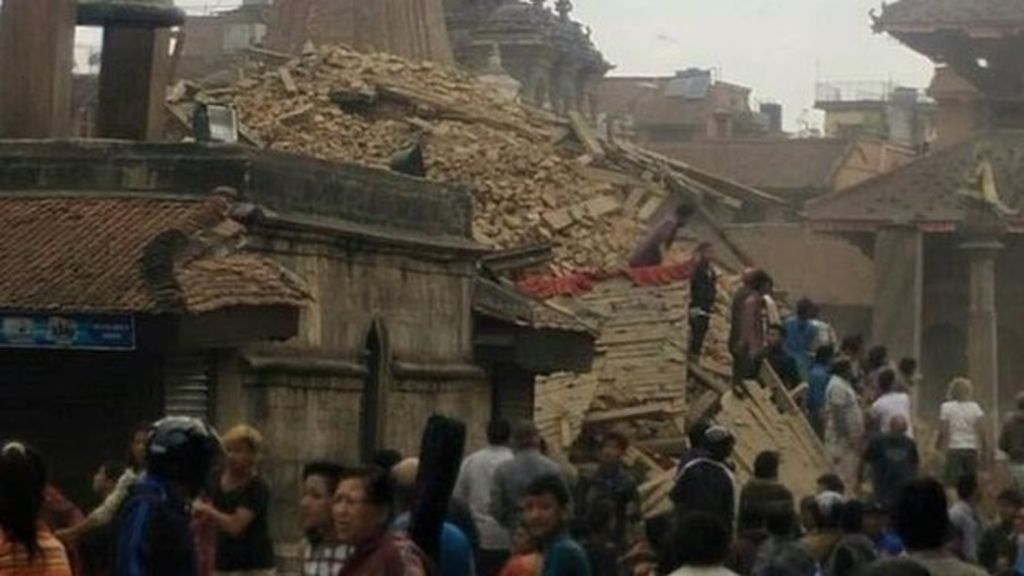} }}%
    \subfloat[(b) DDM]{{\includegraphics[width=1.30in]{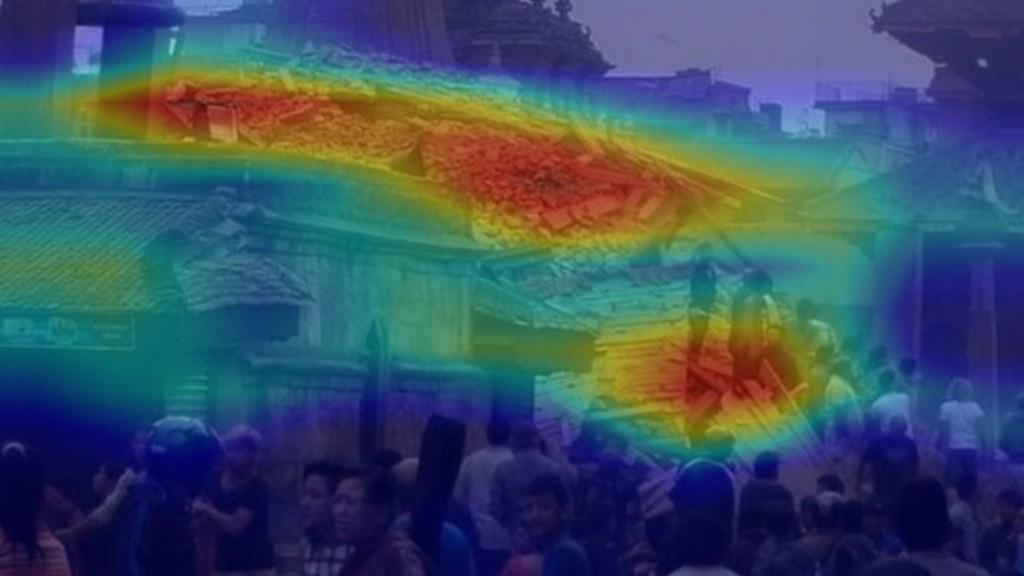} }}%
    \subfloat[(c) DDM+Threshold]{{\includegraphics[width=1.30in]{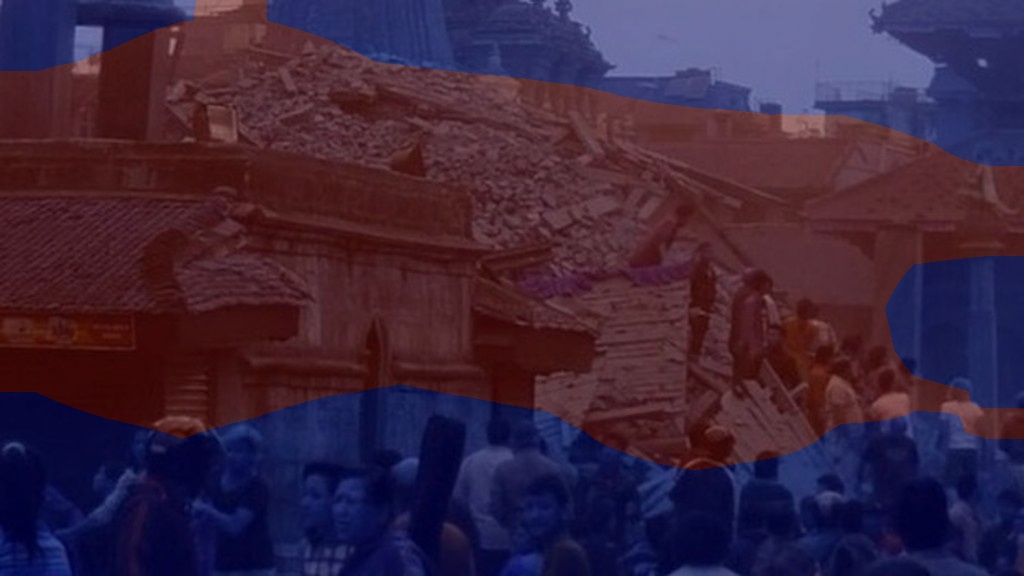} }}%
    \subfloat[(d) Annotator A]{{\includegraphics[width=1.30in]{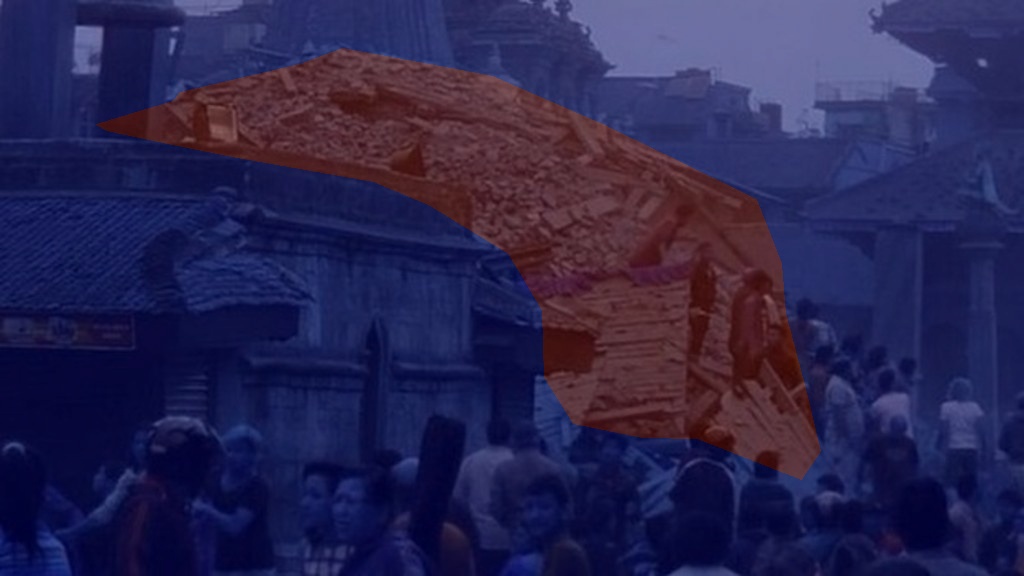} }}%
    \subfloat[(e) Annotator B]{{\includegraphics[width=1.30in]{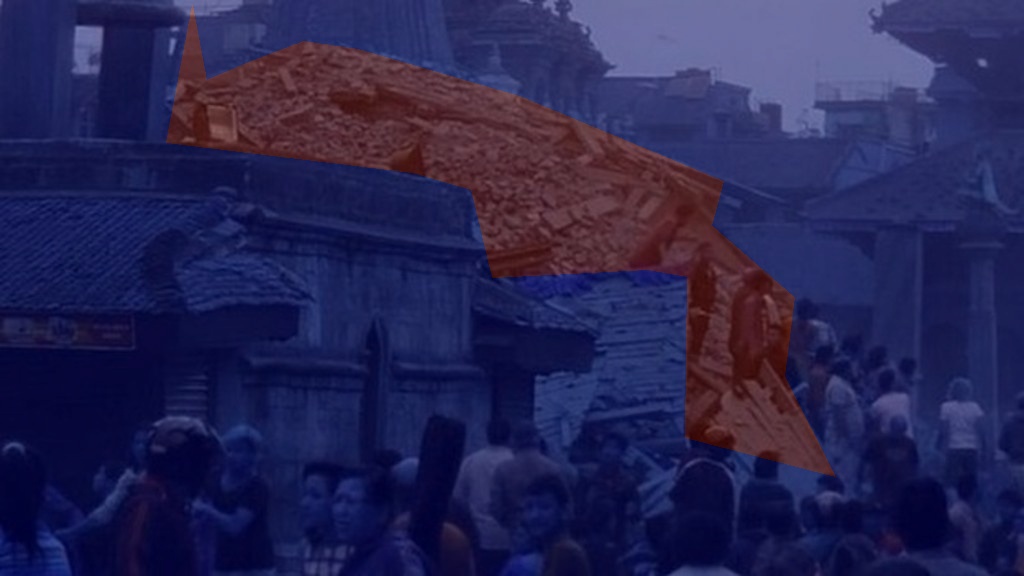} }}
    \caption{Damage Detection Map (DDM) versus human annotations for images 1, 5 and 7 in Table \ref{table:iou}. (a) Original Google image; (b) DDM damage heatmap; (c) Binary representation of the DDM computed with a threshold whose value is 20\% of the max value in the DDM heatmap; (d) Damage marked by Annotator A; (e) Damage marked by Annotator B.  }%
    \label{fig:label_example}%
\end{figure*}

\begin{table*}
\centering
\caption{IOU for 10  Google images. Annotation A and Annotation B are provided independently by two  annotators.  The image heatmaps were transformed to a binary representation by using a threshold equal to 20\% of the max value in the DDM}
\label{table:iou}
\begin{tabular}{|l|l|r|r|r|r|r|r|r|r|r|r|r|}
\hline
 & Image   & 1     & 2     & 3     & 4     & 5     & 6     & 7     & 8     & 9     & 10    & Average \\ \hline
\multirow{2}{*}{IOU} & Annotation A versus  DDM & 0.334 & 0.401 & 0.568 & 0.360 & 0.474 & 0.270 & 0.349 & 0.156 & 0.418 & 0.477 & 0.380 $\pm$ 0.116   \\ 
 &  Annotation B versus DDM & 0.444 & 0.623 & 0.633 & 0.473 & 0.583 &
 0.445 & 0.258 & 0.440 & 0.616 & 0.658 & 0.517 $\pm$ 0.126  \\
 &  Annotation A versus  B &0.677  &0.541 &0.744  & 0.681 &  0.695
 & 0.546  &0.673 &0.237  &0.621  &0.689  &0.610 $\pm$ 0.146  \\ \hline

\end{tabular}
\end{table*}
Conventionally, if the IOU value corresponding to a detected object (marked with a bounding box) is larger than 0.5, the detection/localization of that object is considered to be correct \cite{Everingham:2010}. However, given that the damage is not an object but a concept, we find that the 0.5 threshold is too strict in the context of detecting and localizing damage, especially as the average IOU agreement between annotators is 0.610. If we consider an IOU score of 0.4 as detection, then the GCAM-based approach detects $50\%$ of the damage marked by annotator A, and $90\%$ of the damage marked by annotator B. 
To better understand the results, in Fig. \ref{fig:label_example}, we show the annotations for three sample images, specifically, image 5 for which the IOU agreement with both annotators is above $0.4$, image $7$ for which the IOU agreement with both annotators is smaller than $0.4$, and image 1, for which the agreement with Annotator A is below 0.4 and the agreement with the Annotator B is above 0.4. 
As can be seen, the damage areas marked based on the DDM heatmaps look accurate overall, and in some cases better than the damage areas marked by the annotators, which confirms that identifying damage is a subjective task, and that DDM can be useful in localizing damage and providing visualizations that can help responders gain  trust in the annotations produced by deep learning approaches. 


\subsection{Damage Assessment Value Evaluation}
We calculate the Damage Assessment Value (DAV) as described in Section \ref{DAV}. The DAV values for the sample Google images in Fig. \ref{fig:data_example} are shown below the corresponding heatmap images. As can be seen, images that present a more severe damage scene have higher DAV values, while images with no damage have DAV values very close to zero. 

\begin{figure*}[t]
    \centering
    \subfloat[Mild, DAV=0.328]{{\includegraphics[width=1.30in]{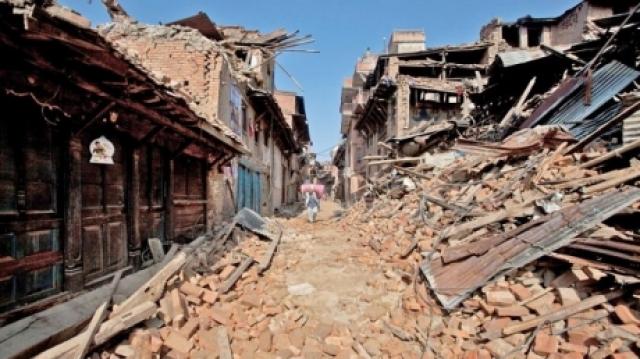} }}%
    \subfloat[None, DAV=0.307]{{\includegraphics[width=1.30in]{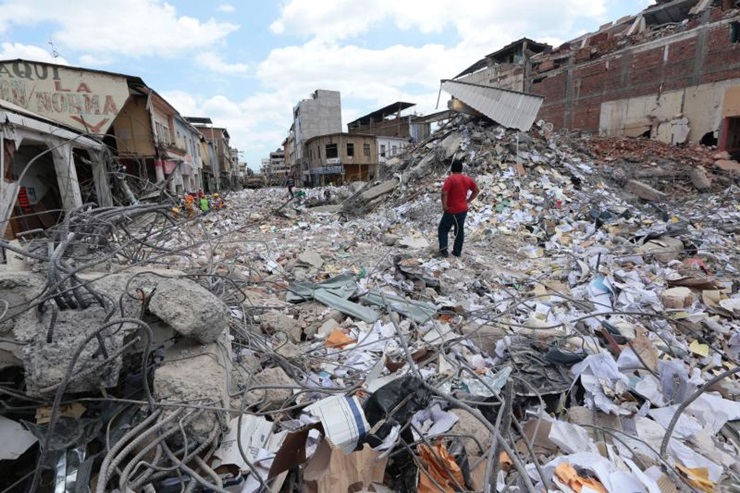} }}%
    \subfloat[None, DAV=0.435]{{\includegraphics[width=1.30in]{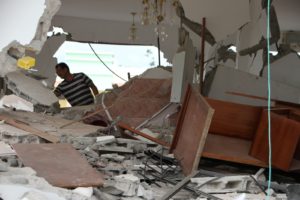} }}%
    \subfloat[Mild, DAV=0.375]{{\includegraphics[width=1.30in]{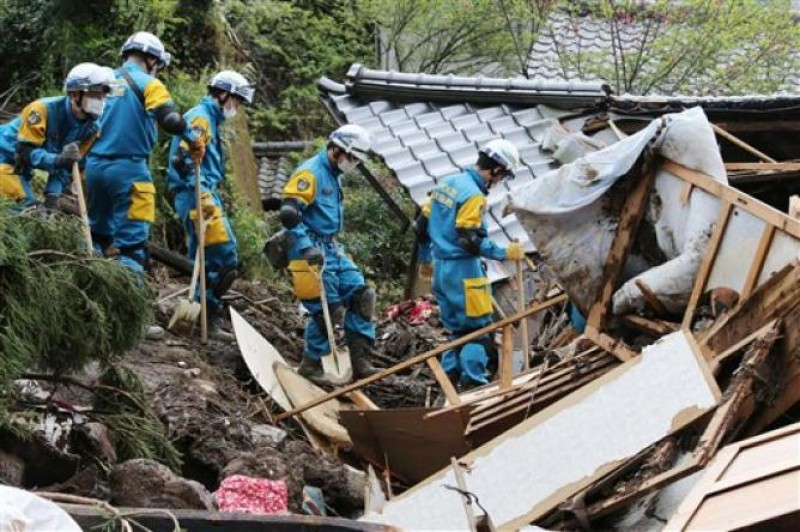} }}%
    \subfloat[Severe, DAV=0.376]{{\includegraphics[width=1.30in]{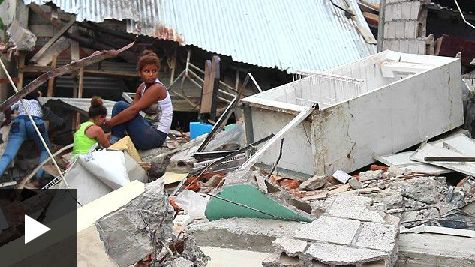} }}%
    \qquad
    \caption{Examples of images mislabeled in the disaster dataset. The original image label  and the DAV score are shown.}%
    \label{fig:label_confusing_example}%
\end{figure*}

\begin{figure*}%
    \centering
    \subfloat[DAV (Nepal Earthquake)]{{\includegraphics[width=3.5cm]{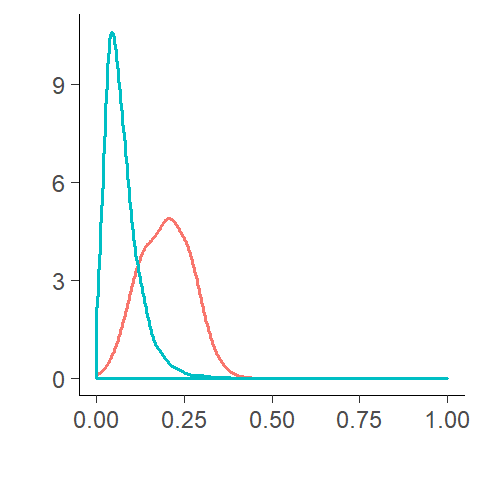} }}%
    \subfloat[DAV (Ecuador Earthquake)]{{\includegraphics[width=3.5cm]{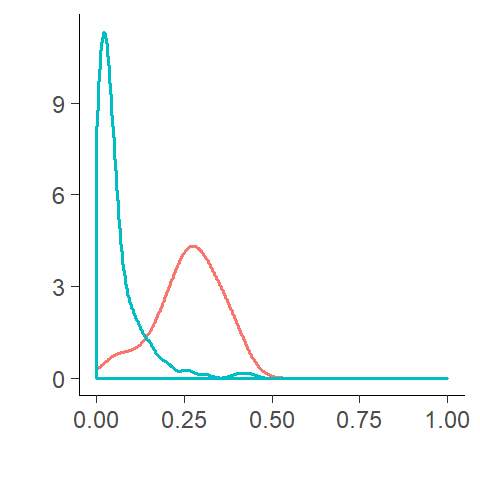} }}%
        \subfloat[DAV (Ruby Typhoon)]{{\includegraphics[width=3.5cm]{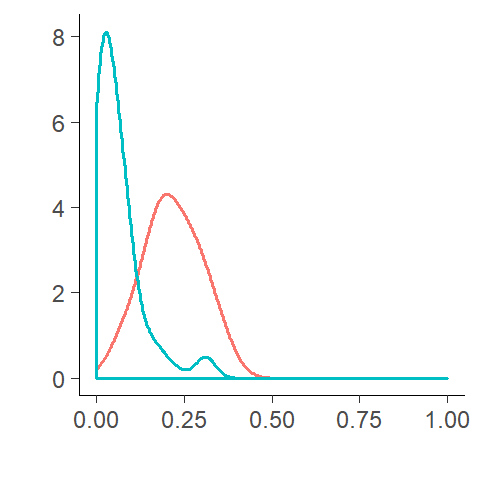} }}%
    \subfloat[DAV (Matthew Hurricane)]{{\includegraphics[width=3.5cm]{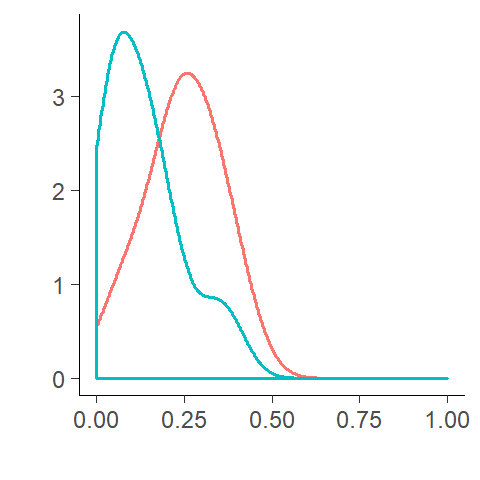} }}%
    \\
    \caption{Smoothed density curves for DAV values. The {\it no damage} class is shown in blue and the {\it damage} class is shown in red. The graphs are based on the test set of each disaster dataset.}
    \label{fig:density}%
\end{figure*}

We also use the disaster datasets shown in Table \ref{table:data_summary} to further evaluate the DAV values. For each disaster, we use the corresponding training set to fine-tune the CNN model to that specific disaster. For this purpose, we combined the original {\it severe} and {\it mild} damage classes into a single {\it damage} class, while the original {\it none} class label is used as {\it no damage} class. We used the CNN model of each disaster to produce DDM heatmaps for all test images from that disaster. Subsequently, we used the heatmaps to generate DAV values for each test image, and generated DAV density plots for the {\it damage} versus {\it no damage} classes, shown in Fig. \ref{fig:density}. As can be seen in the figure, the density plots show clearly differentiable patterns between the two classes, with the {\it damage} class exhibiting higher values overall, as expected. However, there is also overlap in terms of DAV values for {\it damage} and {\it no damage} images. This can be partly explained by the noisiness of the dataset, and the difficulty of the task of separating images in damage severity classes, as also noted in \cite{nguyen2017damage}. To illustrate this claim, in Fig. \ref{fig:label_confusing_example}, we show several images that seem to be mislabeled in the original dataset. 

\subsection{Classification using DAV values}
In this section, we study the usefulness of the DAV values in classifying images into several damage categories. Specifically, we consider the {\it severe}, {\it mild}, and {\it none} categories, as the images in the disasters used in our study are already labeled with these categories. Using the training data, we perform a grid-search to find two threshold values for DAV, denoted by $c_1$ and $c_2$, which minimize the classification error on the training data. Using these thresholds, we design a simple classifier as follows: test images with DAV values smaller than $c_1$ will be classified as {\it none}, those with DAV values in between $c_1 \mbox{ and } c_2$ will be classified as {\it mild}, and finally those with DAV values greater than $c_2$ will be classified as {\it severe}. The classification results of our simple classifier (averaged over five independent runs), together with the average results of a three-class CNN model (trained on the corresponding training set of each disaster in each run), are reported in Table \ref{table:classification_summary}.
 The classification results of the simple classifier based on DAV are similar to those of the CNN model for Ecuador Earthquake and Matthew Hurricane (i.e., the results are not statistically different based on a t-test with \(p \leq 0.05\)). While the CNN accuracy looks better overall, the results indicate that DAV can capture severity damage (on a continuous scale), and it can help produce simple and interpretable classifiers. 

\begin{table}
\centering
\caption{Classification accuracy for each disaster dataset. The first row shows the average accuracy of our simple DAV-based classifiers (over 5 independent runs), together with standard deviation. The second row shows the average accuracy of the three-class CNN models. The third row shows the results published in \cite{NguyenAJSIM16}, which used the same datasets.}
\label{table:classification_summary}
\scriptsize
\begin{tabular}{|l|l|l|l|l|}
\hline
Method  & \multicolumn{1}{l|}{\begin{tabular}[c]{@{}l@{}}Nepal \\ Earthquake\end{tabular}} & \multicolumn{1}{l|}{\begin{tabular}[c]{@{}l@{}}Ecuador \\ Earthquake\end{tabular}} & \multicolumn{1}{l|}{\begin{tabular}[c]{@{}l@{}}Ruby \\ Typhoon\end{tabular}} & \multicolumn{1}{l|}{\begin{tabular}[c]{@{}l@{}}Matthew \\ Hurricane\end{tabular}} \\ \hline
DAV    & 0.801$\pm$0.013  & 0.886$\pm$0.018 &0.793$\pm$0.017   & 0.533$\pm$0.043  \\ \hline
VGG19    & 0.851$\pm$0.003  & 0.901$\pm$0.013  & 0.852$\pm$0.018  & 0.603$\pm$0.151  \\ \hline
VGG16 \cite{NguyenAJSIM16} & 0.840 & 0.870 & 0.810 & 0.740 \\ \hline
\end{tabular}
\end{table}

\section{Related Work}
\label{related}
Social media data has been shown to have significant value in disaster response  \cite{castillo2016big,Meier:2015,palen2016crisis}. Many machine learning approaches \cite{ashktorab2014tweedr,imran2015processing,hongmin:JCCM}, including deep learning approaches \cite{CarageaST16,NguyenAJSIM16}, have been proposed and used to help identify and prioritize useful {\it textual} information (e.g., tweets) in social media. Some works have focused specifically on identifying situational awareness information \cite{sen2015extracting,huang2015geographic}, including information related to damage assessment \cite{kryvasheyeu2016rapid,yuan2018feasibility,resch2017combining,DBLP:journals/corr/abs-1802-02631}.

Despite the extensive use of machine learning tools for analyzing social media text data posted during disaster events, there is not much work on analyzing social media images posted by eyewitnesses of a disaster. One pioneering work in this area \cite{nguyen2017damage}, used trained CNN models, specifically, VGG16 networks fine-tuned on the disaster image datasets that are also used in our work (see Table \ref{table:data_summary}), and showed that the CNN models perform better than standard techniques based on bags-of-visual-words. 
Our CNN results, using VGG19 fine-tuned on the same disaster image datasets, are similar to those reported in \cite{nguyen2017damage}, except for Matthew Hurricane, for which the dataset is relatively small and the model can't be trained well. However, as opposed to \cite{nguyen2017damage}, where the focus is on classifying images into three damage categories, we go one step further and use the Grad-CAM approach \cite{selvaraju2016grad} to localize damage, with the goal of improving the trust of the response teams in the predictions of the model. Furthermore, we produce a continuous severity score, as a way to quantify damage.  

Other prior works focused on image-based disaster damage assessment use aerial or satellite images, e.g. \cite{xie2016crowdsourcing,gueguen2015large}. 
Compared to such works, which use more expensive imagery, we focus on the use of social media images, which are readily available during disasters, together with interpretability approaches, i.e., Grad-CAM, to produce a damage map and a damage severity score for each image. 

Similar to us, Nia and Mori \cite{nia2017building}  use ground-level images collected using Google to assess building damage. Their model consists of three different CNN networks (fine-tuned with  raw images, color-masked or binary-masked images, respectively) to extract features predictive of damage. Subsequently, a regression model is used with the extracted  features to predict the severity of the damage on a continuous scale. Compared to \cite{nia2017building}, we use the features identified at the last convolutional layer of the CNN network to build a detection map, and use the map to produce a numeric damage severity score. While CAM-type approaches have been used to explain model predictions in many other application domains, to the best of our knowledge, they have not been used to locate damage and assess  damage severity in prior work. 


\section{Conclusion}
\label{conc}
Given the large number of social media images posted by eyewitnesses of disasters, we proposed an approach for detecting and localizing disaster damage at low cost. Our approach is built on top of a fine-tuned VGG19 model, and utilizes the Grad-CAM approach to produce a DDM heatmap. Furthermore, the DDM is used to calculate a DAV score for each image. This scoring is performed on a continuous scale and can be used to assess the severity of the damage. The DAV score, together with the DDM heatmap, can be used to identify and prioritize useful information for disaster response, while providing visual  explanations for the suggestions made to increase the trust in the computational models. Quantitative and qualitative evaluations of DDM and DAV components show the feasibility of our proposed approach. 

As part of future work, we will compare the Grad-CAM approach used to produce the damage heatmaps with its newer variant Grad-CAM++ \cite{Chattopadhyay}, which the authors claim to produce heatmaps which better cover the objects of interest as opposed to just identifying the most prominent object features. 
Also of interest is the applicability of the proposed approach to estimate the global damage produced by a disaster based on aggregating the DAV values from individual images. Finally, geo-tagging images would enable disaster response teams not only to identify damage, but also to find its physical location. 

\bibliographystyle{IEEEtran.bst}
\bibliography{egbib}

\end{document}